\documentclass[letterpaper]{article} % DO NOT CHANGE THIS
\usepackage{aaai2026}  % DO NOT CHANGE THIS
\usepackage{times}  % DO NOT CHANGE THIS
\usepackage{helvet}  % DO NOT CHANGE THIS
\usepackage{courier}  % DO NOT CHANGE THIS
\usepackage[hyphens]{url}  % DO NOT CHANGE THIS
\usepackage{graphicx} % DO NOT CHANGE THIS
\usepackage{natbib}  % DO NOT CHANGE THIS AND DO NOT ADD ANY OPTIONS TO IT
\usepackage{caption} % DO NOT CHANGE THIS AND DO NOT ADD ANY OPTIONS TO IT
\usepackage{algorithm}
\usepackage{algorithmic}
\usepackage{cite}
\usepackage{amsmath,amssymb,amsfonts}
\usepackage{algorithmic}
\usepackage{graphicx}
\usepackage{textcomp}
\usepackage{xcolor}
\usepackage{cite}
\usepackage{csquotes}
\usepackage[inkscapelatex=false]{svg}
\usepackage{array,ragged2e,booktabs,nicematrix}
\usepackage{subcaption}
\usepackage{adjustbox}
\usepackage{comment}
\usepackage{todonotes}
\usepackage{dsfont}
\usepackage{amssymb}
\newcommand{\qed}{\null\nobreak\hfill\ensuremath{\square}}

\usepackage{newfloat}
\usepackage{listings}
\DeclareCaptionStyle{ruled}{labelfont=normalfont,labelsep=colon,strut=off} % DO NOT CHANGE THIS
\floatstyle{ruled}
\newfloat{listing}{tb}{lst}{}
\floatname{listing}{Listing}
\newcommand{\R}[0]{\mathbb{R}}
\newcommand{\Rd}[0]{\mathbb{R}^d}
\newcommand{\ex}[0]{\mathbb{E}}
\newcommand{\var}[0]{\mathbb{V}}
\DeclareMathOperator{\argmin}{\text{arg\,min}}
\newtheorem{theorem}{Theorem}
\newcommand{\upar}{{\Large\textbf{\textuparrow}}}
\newcommand{\doar}{{\Large\textbf{\textdownarrow}}}
\newcommand{\updir}{\upar \hspace{0.15cm}}
\newcommand{\downdir}{\doar \hspace{0.15cm}}

\title{
The Directed Prediction Change - Efficient and Trustworthy Fidelity Assessment\\for Local Feature Attribution Methods
}
\author {
    First Author Name\textsuperscript{\rm 1,\rm 2},
    Second Author Name\textsuperscript{\rm 2},
    Third Author Name\textsuperscript{\rm 1}
}
\author{
Kevin Iselborn$^{\ast}$,
David Dembinsky$^{\ast}$,
Adriano Lucieri,
Andreas Dengel
}
\affiliations {
    \emph{German Research Center for Artificial Intelligence (DFKI) \&}\\ 
    \emph{Department of Computer Science, }\emph{RPTU University Kaiserslautern-Landau}\\
    67663 Kaiserslautern, Germany \\
    firstname.lastname@dfki.de
}

\begin{document}

\maketitle

\begingroup
\renewcommand\thefootnote{\fnsymbol{footnote}}
\footnotetext[1]{These authors contributed equally.}
\endgroup

\begin{abstract}
    The utility of an explanation method critically depends on its fidelity to the underlying machine learning model.
    Especially in high-stakes medical settings, clinicians and regulators require explanations that faithfully reflect the model's decision process.
    Existing fidelity metrics such as \textit{Infidelity} rely on Monte Carlo approximation, which demands numerous model evaluations and introduces uncertainty due to random sampling.
    This work proposes a novel metric for evaluating the fidelity of local feature attribution methods by modifying the existing \textit{Prediction Change} (\textit{PC}) metric within the Guided Perturbation Experiment.
    By incorporating the direction of both perturbation and attribution, the proposed \textit{Directed Prediction Change} (\textit{DPC}) metric achieves an almost tenfold speedup and eliminates randomness, resulting in a deterministic and trustworthy evaluation procedure that measures the same property as local \textit{Infidelity}.
    \textit{DPC} is evaluated on two datasets (skin lesion images and financial tabular data), two black-box models, seven explanation algorithms, and a wide range of hyperparameters.
    Across $4\,744$ distinct explanations, the results demonstrate that \textit{DPC}, together with \textit{PC}, enables a holistic and computationally efficient evaluation of both baseline-oriented and local feature attribution methods, while providing deterministic and reproducible outcomes. 
\end{abstract}

\section{Introduction}%
\begin{figure}[t]
    \centering
    \begin{subfigure}{\linewidth}
        \centering
        \includesvg[width=\linewidth]{images/infidelity_confidence_ISIC-Inception-0.2_clean.svg}
        \caption{\downdir Infidelity mean and approximation std. for all tested methods}
        \label{fig:intro_infidelity_confidence_and_runtime:uncertainty_image}
    \end{subfigure}\\
    \begin{subfigure}{\linewidth}
        \centering
        \begin{adjustbox}{width=\linewidth}
            \begin{NiceTabular}{ccccc}
                \toprule
                        \Block[c,c]{1-1}{} & \Block[c,c]{1-1}{Number of\\Model Evaluations} & \Block[c,c]{1-3}{Runtime [s] \\ Mean / Median $\pm$ std} & & \\\midrule
                        \Block[c,l]{1-1}{Infidelity} & \Block[c,c]{1-1}{$640$} & \Block[c,c]{1-1}{$4924.83 / 5714.21$} &  \Block[c,l]{1-1}{$\pm$}&  \Block[c,r]{1-1}{$1170.52$}  \\\cmidrule{2-5}
                        \Block[c,l]{1-1}{DPC (ours)} & \Block[c,c]{1-1}{$\mathbf{40}$} & \Block[c,c]{1-1}{$\mathbf{593.54} / \mathbf{576.67}$} &  \Block[c,l]{1-1}{$\pm$} &  \Block[c,r]{1-1}{$42.35$} \\\midrule
                        \Block[c,l]{1-1}{Speedup} & \Block[c,c]{1-1}{} & \Block[c,c]{1-1}{$\mathbf{8.30} / \mathbf{9.91}$} & &  \\
                \bottomrule
            \end{NiceTabular}
        \end{adjustbox}
        \caption{Running time for an A100-80GB GPU when evaluating with \textit{Infidelity} and \textit{DPC} (the proposed metric) on skin lesion image data.}
        \label{fig:intro_infidelity_confidence_and_runtime:runtime_table}
    \end{subfigure}
    
    \caption{\textbf{(a)} The Monte Carlo sampling used by \textit{Infidelity} exhibits high variance, requiring many samples for reliable FA evaluation (i.e., low standard deviation). 
\textbf{(b)} The proposed \textit{DPC} metric achieves an almost tenfold speedup while providing a deterministic, and hence trustworthy, evaluation.}
    \label{fig:intro_infidelity_confidence_and_runtime}
\end{figure}
\noindent
The application of artificial intelligence (AI) in high-risk domains such as healthcare bears great potential, yet also entails substantial risks that must be mitigated through regulations demanding trustworthiness.
Beyond thorough performance evaluation, explainable AI (XAI) methods aim to enhance the transparency of modern black-box models and facilitate their integration into clinical practice.
The rapid development of AI has led to a vast variety of XAI techniques~\cite{arrieta2020explainable, yang2023survey}, making it increasingly difficult to select an appropriate method. 
Moreover, explanation algorithms in general, and feature attribution (FA) methods in particular, often involve numerous hyperparameters.
Consequently, an extensive evaluation across many configurations is required to meaningfully assess the performance of any single algorithm.

Existing analyses of explanation algorithms often rely on anecdotal evidence, that is, the qualitative inspection of exemplary explanations \cite{nauta2023anecdotal, dembinsky2025unifying}.
While such anecdotal evidence can provide users with an intuition about explanation behavior, it is inherently subjective \cite{bucinca2020proxy, nauta2023anecdotal, dembinsky2025unifying}.
Furthermore, human evaluation is prone to confirmation bias, focusing predominantly on plausibility while neglecting aspects such as the faithfulness of the explanation to the underlying decision process \cite{doshi2017towards}.
To enable systematic and trustworthy evaluation, quantitative and functionally grounded evaluation metrics are therefore required to provide proxies for measuring explanation quality.

Recently, the \emph{eValuation of XAI} (\emph{VXAI}) framework extensively categorized such metrics \cite{dembinsky2025unifying}.
Among the described desiderata (i.e., desirable atomic properties of explanations), fidelity is particularly important: without fidelity, even explanations that score well on other desiderata cannot provide meaningful insight into the model \cite{dembinsky2025unifying} and are thus unsuitable for high-risk applications such as medicine or finance.
The most common metrics in this category are based on the \emph{Unguided} and \emph{Guided Perturbation Experiment}, which evaluate fidelity through \emph{input intervention} by perturbing input features and comparing the resulting change in model prediction with the effect predicted by the explanation \cite{dembinsky2025unifying}.

A widely used metric performing the \emph{Unguided Perturbation Experiment} to compare FA methods is \textit{Infidelity}.
\textit{Infidelity} estimates the expected agreement between an explanation and the corresponding change in model prediction under small perturbations using Monte Carlo sampling. 
However, ensuring a trustworthy evaluation of FA methods requires a large number of Monte Carlo samples, and thus repeated model evaluations, resulting in high computational cost (see Figure \ref{fig:intro_infidelity_confidence_and_runtime:uncertainty_image} and Section \ref{sec:experiments:results:dpc_efficiency}).
This makes the scalable and trustworthy selection of FA methods impractical.

To address this limitation, this paper proposes a modification to the closely related \emph{Guided Perturbation Experiment}.
This work demonstrates that while the original Guided Perturbation Experiment is suitable only for FA methods that compare an input to a reference baseline (i.e., \textit{baseline-oriented} methods), incorporating the direction of perturbation and attribution yields a metric that also applies to local FA methods.
The resulting \emph{Directed Prediction Change (DPC)} is a novel, computationally efficient, and deterministic metric that achieves an almost tenfold median speedup (see Figure \ref{fig:intro_infidelity_confidence_and_runtime:runtime_table}) compared to \textit{Infidelity} \cite{yeh2019infidelity}, which, to the best of our knowledge, is the only other metric targeting these explanation approaches. 

\section{Background}
\label{sec:related}

\subsection{Feature attribution explanations}
A FA method is a mapping that assigns each feature of an input a significance score, indicating which parts of the input were relevant for the predictor's outcome. 
Let $y$ be the target class of interest, scored by a black-box model $f$.
Furthermore, let $s_f^{y}: \R^d \to \R$ denote the scalar scoring function for class $y$ and model $f$.
A FA method is a function $\mathcal{A}_{f}^{y}:\R^d \to \R^d$
that assigns a relevance score $a_i := \mathcal{A}_{f}^{y}(x)_i$ to each feature $i \in \{1, \dots, d\}$ of an input $x\in\R^d$.  

Positive values $(a_i>0) $ provide supporting evidence for the score of $f$, negative values $(a_i<0) $ provide contradicting evidence.
The magnitude $|a_i|$ reflects the strength of the contribution.
The interpretation of supporting or contradicting evidence depends on whether the FA method provides local or baseline-oriented explanations\footnote{\citeauthor{ancona2017towards} refer to these approaches as \emph{local} and \emph{global attributions} \cite{ancona2017towards}. However, we find these terms misleading with respect to \emph{local} and \emph{global model explanations} as defined by other frameworks \cite{speith2022review, molnar2020interpretable}.}.

For \emph{local} methods, in a sufficiently small neighborhood around the data point, increasing $x_i$ is expected to increase (support) or decrease (contradict) the score \cite{simonyan2013deep, springenberg2014striving, ribeiro2016why}. 

For \emph{baseline-oriented} methods, support (contradiction) indicates that the difference between the data point and the baseline in feature $x_i$ increases (decreases) the score relative to the baseline \cite{sundararajan2017axiomatic, lundberg2017shap}.
Baseline-oriented methods are therefore inherently contrastive \cite{sundararajan2017axiomatic} and follow a different paradigm than local FA methods.

This analysis considers both local and baseline-oriented methods.
To cover a wide range of approaches and include widely adopted techniques, we use the following methods:
\begin{itemize}
    \item \textbf{Vanilla Gradient} (Grad) \cite{simonyan2013deep}:
    \begin{equation}
        \mathbf{Grad}_f^y(x) = \nabla_x s_f^y(x)
    \end{equation}
    \item \textbf{Guided Backpropagation} (GB) \cite{springenberg2014striving}: Modifies the backpropagation rules of ReLU activations as follows 
    \begin{equation}
        \frac{\partial s_f^y(x)}{\partial g(x)} \;:=\; R  \cdot \mathds{1}[g(x)>0] \cdot\mathds{1}[R>0],
    \end{equation}
    where $g(x)$ denotes the input to a ReLU activation function, and $R:= \left(\partial s_f^y(x)\right)/\left(\partial \mathrm{ReLU}\big(g(x)\big)\right)$ is the upstream gradient at that node. 
    \item \textbf{SmoothGrad} (SG) \cite{smilkov2017smoothgrad} and \textbf{VarGrad} (VG) \cite{adebayo2018local}: Compute the average or the variance of the FAs over noisy inputs, respectively.
    \begin{equation}
        \mathbf{SG}_f^y(x) = \ex_{X\sim\mathcal{N}(x,\sigma I)}\!\big[ \nabla_X s_f^y(X) \big]
    \end{equation}
    \begin{equation}
        \mathbf{VG}_f^y(x) = \var_{X\sim\mathcal{N}(x,\sigma I)}\!\big[ \nabla_X s_f^y(X) \big]
    \end{equation}
    \item \textbf{Integrated Gradients} (IG) \cite{sundararajan2017axiomatic}: Compute the path integral of the gradients between a baseline $x_0$ and the input to be explained.
    \begin{equation}
        \mathbf{IG}_{f}^y(x) = (x-x_0) \int_0^1 \nabla_x s_f^y\big(x_0 + \alpha(x-x_0)\big)\mathrm{d}\alpha, 
    \end{equation}
    \item \textbf{LIME} \cite{ribeiro2016why}: Trains a linear model $g\in G$ to locally approximate the model prediction and then uses the model as an explanation.
    \begin{equation}
        \mathbf{LIME}_f^y(x) := \argmin_{g\in G} \mathcal{L}^y(f,g,x) + \Omega(g),
    \end{equation}
    where $\mathcal{L}$ is a loss function and $\Omega$ is a measure of complexity for the resulting explanation. This results in a large number of hyperparameters for LIME. 
    
    \item \textbf{DeepLiftSHAP} \cite{lundberg2017shap}: Efficiently approximates Shapely Values (SV) by adapting the backpropagation rules of DeepLift. \small
     \begin{equation}
         \textbf{SV}_f^y(x)_i:=\scalebox{0.8}{$  \sum\limits_{S\subseteq\{1,\dots,d \}\setminus\{i\}} \frac{|S|!(d-|S|-1)!}{d!}\left(s_f^y\left(x_{S\cup\{i\}}\right) - s_f^y(x_{S}) \right)$}
     \end{equation} \normalsize
\end{itemize}

In addition to these FA methods, we also employ two model-agnostic reference methods, which we expect to be exceeded by the FA methods.
We use importance values drawn from a standard gaussian distribution as \emph{random explanations} for both considered data modalities.
On image data, we additionally apply an \emph{edge detection algorithm}.

A wide variety of hyperparameters is considered for all methods, with further details provided in the Appendix. 

\subsection{Evaluation of XAI}

\subsubsection{Unguided Perturbation Fidelity}
Let $\pi: \R^d \to \R^d$ denote a perturbation function sampled from a distribution $\Pi$, and let $s_f^{y}: \R^d \to \R$ be the scoring function for class $y$ and model $f$.
For a FA method $\mathcal{A}_{f}^{y}$, the \emph{Infidelity} metric  \cite{yeh2019infidelity} on input $x \in \R^d$ is defined as 
\begin{align}
    P_{f}^{y}(\pi, x) :=&(x - \pi(x))^T \mathcal{A}_{f}^{y}(x) \label{eq:infid_pert_effect}\\ % \notag
    S_{f}^{y}(\pi, x) :=&s_f^{y}(x) - s_f^{y}\big(\pi(x)\big)\\ %\notag
    \mathbf{Inf}_{f}^{y}(x) 
    := &\ex_{\pi \sim \Pi} \left[
        \big(P_{f}^{y}(\pi, x) - S_{f}^{y}(\pi, x)\big)^2
    \right].
\label{eq:infid_main_equation}
\end{align}
\textit{Infidelity} represents the expected mean squared error between the predicted perturbation effect $P$ and the actual score change $S$.
The choice of perturbation distribution $\Pi$ is therefore crucial, as suitable choices enable the evaluation of both baseline-oriented and local FA methods.
We consider adding gaussian noise with $\sigma = 0.2$ to measure local \textit{Infidelity}. %and replacement with the zero baseline for baseline-oriented \textit{Infidelity}.
Smaller values indicate a smaller mismatch and therefore higher fidelity (\doar).

\subsubsection{Guided Perturbation Fidelity}
In guided approaches, perturbations are applied sequentially to features according to their attribution ranking rather than being sampled randomly.  
At each step $t$, the \textit{Prediction Change (PC)} is defined as the drop in class score when the next feature is removed:
\begin{equation}
    \mathrm{PC}^y_t(x) \;=\; s_f^{y}\big(\pi_t(x)\big) - s_f^{y}\big(\pi_{t-1}(x)\big),
\end{equation}
where $\pi_t: \R^d \to \R^d$ denotes the perturbation operator after $t$ steps.  
Faithful explanations should cause large prediction drops (i.e., negative or low \emph{PC}) early when features are removed in descending order of importance (MoRF), and delayed drops when removed in the reverse order (LeRF).  
This behavior is typically quantified by the \emph{Area Under the Perturbation Curve} (AUPC).  
To capture both perspectives, we report the weighted \emph{Area Between Perturbation Curves} (ABPC) \cite{simicPerturbationEffectMetric2022}, which measures the gap between LeRF and MoRF curves, weighting early changes more strongly.
For the resulting measure, larger values indicate a better FA (\upar).

\section{The \textit{Directed Prediction Change} (\textit{DPC})}
\label{sec:new_metric}

\subsection{\textit{Prediction Change} (\textit{PC}) is insufficient for the evaluation of local FA methods} \label{sec:new_metric:pc_analysis}

\begin{figure}[t]
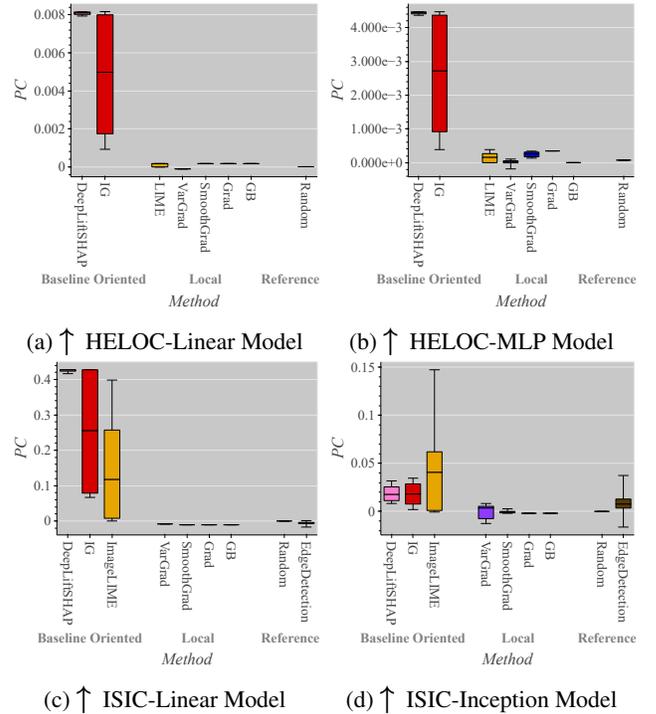

    \centering
    \begin{subfigure}{0.49\linewidth}
        \centering
        \includesvg[width=\linewidth]{images/HELOC-Linear_MBI=True_Method_PC_0_clean.svg}
        \caption{\updir HELOC-Linear Model}
    \end{subfigure}
    \begin{subfigure}{0.49\linewidth}
        \centering
        \includesvg[width=\linewidth]{images/HELOC-MLP_MBI=True_Method_PC_0_clean.svg}
        \caption{\updir HELOC-MLP Model}
    \end{subfigure}\\
    \begin{subfigure}{0.49\linewidth}
        \centering
        \includesvg[width=\linewidth]{images/ISIC-Linear_MBI=True_Method_PC_0_clean.svg}
        \caption{\updir ISIC-Linear Model}
    \end{subfigure}
    \begin{subfigure}{0.49\linewidth}
        \centering
        \includesvg[width=\linewidth]{images/ISIC-Inception_MBI=True_Method_PC_0_clean.svg}
        \caption{\updir ISIC-Inception Model}
    \end{subfigure}\\
    \caption{
        Overview of \textit{Prediction Change} (\textit{PC}) results for all considered models.
        The \textit{PC} ranks baseline-oriented FA methods higher than other methods, whereas the \textit{Gradient} (an optimal local FA method for linear models) performs comparably to random attribution.
    }
    \label{fig:method_pc}
\end{figure}

To motivate our novel metric, we first revisit the Sensitivity-$N$ property introduced by \citet{anconaGradientBasedAttributionMethods2019}.
A FA method satisfies Sensitivity-$N$ for an $x \in \Rd$ if and only if, for a baseline $x'\in \R^d$, any subset $S$ of the feature set $X = \{1, \dots, d\}$ with $|S| = N$ satisfies  
\begin{equation} \label{eq:sensitivity_N}
    \sum_{i \in S} \mathcal{A}_{f}^y(x)_i = s_f(x) - s_f(x_{X/S};x'),
\end{equation}
where $x_{X/S};x'$ denotes replacing the values in $x$ by those from $x'$ for the features in $S$. 

Satisfying Sensitivity-$N$ for all $N$ is a sufficient condition for baseline-oriented FA methods.
Sensitivity-1 implies that each feature's attribution exactly corresponds to the difference in model prediction, thereby fulfilling the condition of a baseline-oriented FA.
Analogously, $N > 1$ extends this to combinations of multiple features.
\citet{anconaGradientBasedAttributionMethods2019} show that several FA methods, including \textit{Integrated Gradients} and \textit{DeepLift}, satisfy Sensitivity-$N$ for all $N$ and $x\in \Rd$ if applied to linear models.
These can therefore be regarded as theoretically confirmed baseline-oriented FA methods. 

Under mild assumptions (which hold for the FA methods analyzed by \citet{anconaGradientBasedAttributionMethods2019} in linear settings), Sensitivity-$N$ is also a sufficient condition for optimality according to the \textit{PC} metric.
We formalize this in the following theorem, with the proof provided in the Appendix:

\begin{theorem}
    Let $x$ be the considered data point, $x' \in \Rd$ be a baseline, $\mathcal{A}$ a FA method with consistent rankings on all perturbation paths between $x$ and $x'$, and $s_f^y$ the scoring function for model $f$ and class $y$. 
    
    If $\mathcal{A}$ satisfies Sensitivity-$N$ for $s_f^y$ on all perturbation paths between $x$ and $x'$ for all $1 \leq N \leq d$, then $\mathcal{A}$ is optimal under evaluation by the Prediction Change (\textit{PC}) on $x$.
\end{theorem}

This shows that the baseline-oriented FA methods analyzed by \citet{anconaGradientBasedAttributionMethods2019} are rated as optimal by \textit{PC}.
In this work, we consider \textit{Integrated Gradients} and \textit{DeepLiftSHAP}, which have been shown to satisfy Sensitivity-$N$ and are thus optimal under \textit{PC} for linear models\footnote{\textit{DeepLiftSHAP} inherits the Sensitivity-$N$ property from \textit{DeepLift} through the use of the mean baseline.}.

While this theoretical analysis does not extend to nonlinear models, we experimentally assess the FA methods across all hyperparameters described in the Appendix in the boxplots presented in Figure \ref{fig:method_pc}. 
As is done for all boxplots in this work, whiskers indicate maximal and minimal values to highlight the best and worst hyperparameter configurations observed.
It is observed, that for all considered models (linear and nonlinear), baseline-oriented FA methods are rated superior by \textit{PC}. 

\citet{anconaGradientBasedAttributionMethods2019} further show that many baseline-oriented methods can be expressed using input multiplication. Therefore they can be reformulated as $(x - x') \cdot z$ for some $z \in \Rd$ and baseline $x' \in \Rd$, where $\cdot$ denotes the Hadamard product.
Retaining only $z$ (and thus \enquote{removing the input multiplication}) then transforms a method into a pseudo-local FA method \cite{anconaGradientBasedAttributionMethods2019}, enabling a more detailed analysis of whether a metric favors local or baseline-oriented attributions.

\begin{figure}[t]
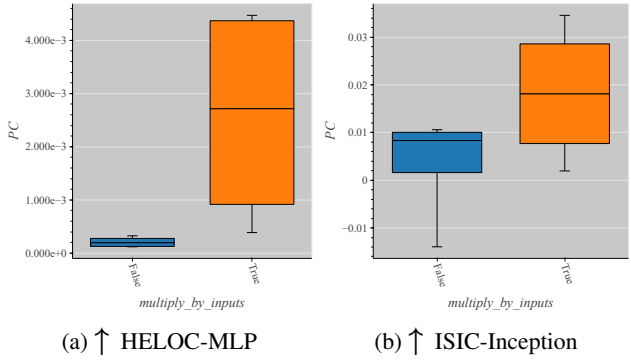

    \centering
    \begin{subfigure}{0.49\linewidth}
        \centering
        \includesvg[width=\linewidth]{images/HELOC-MLP_MBI-IG_PC_0_clean.svg}
        \caption{\updir HELOC-MLP}
    \end{subfigure}
    \begin{subfigure}{0.49\linewidth}
        \centering
        \includesvg[width=\linewidth]{images/ISIC-Inception_MBI-IG_PC_0_clean.svg}
        \caption{\updir ISIC-Inception}
    \end{subfigure}\\
    \caption{
        \textit{Prediction Change} (\textit{PC}) evaluation of \textit{Integrated Gradients} for nonlinear models when baseline-oriented methods are converted into pseudo-local variants ($multiply\_by\_inputs=$ False) following \citet{anconaGradientBasedAttributionMethods2019}.
        As with the Sensitivity-$N$ property, the pseudo-local variants are consistently rated lower by \textit{PC} than their baseline-oriented counterparts.
    }
    \label{fig:mbi_pc}
\end{figure}

Figure \ref{fig:mbi_pc} presents the results for nonlinear models using \textit{Integrated Gradients}.
The pseudo-local variants are consistently rated worse than their baseline-oriented counterparts.
Together, both experiments show that for both linear and nonlinear models, baseline-oriented FA methods are systematically favored by \textit{PC}.

However, the gradient is ranked only marginally better than random attributions, which is problematic for linear models since it represents a well-established correct explanation of model behavior (see, e.g., \citet{molnar2020interpretable}). 
Furthermore, local FA methods are by design better suited to capture local model behavior than baseline-oriented methods.
Accordingly, \textit{PC} is not suitable when the goal is to evaluate explanations of local model behavior. 

\subsection{An intuition for the \textit{Directed Prediction Change} metric}
We now demonstrate how the existing \textit{PC} metric can be adapted to effectively evaluate local FA methods. 

Recall that the \emph{class score} is given by $s_f^y(x)$, where a higher value represents a higher likelihood for class $y$.  
For a local FA method, the \emph{attribution score} $a_i(x)$ for feature $i$ is defined such that a positive value indicates that increasing $x_i$ supports the score of class $y$, whereas a negative value indicates that decreasing $x_i$ reduces $s_f^y(x)$.   

We now analyze how the \textit{PC} behaves when evaluating a local FA method.
In a Guided Perturbation Experiment, attributions must rank features such that at each step the feature causing the minimum (MoRF order) or maximum (LeRF order) \textit{PC}  is selected.
However, we observe that \textit{PC} can only recognize this correctly if the attribution and perturbation directions align.

\begin{figure}[t]
    \centering
    \includegraphics[width=\linewidth]{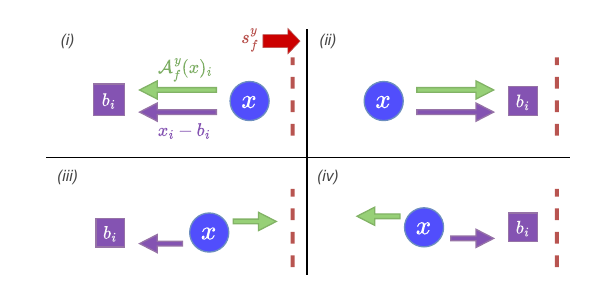}
    \caption{For a perturbation step $t$ in which feature $i$ of data point $x$ is replaced with a baseline value $b$, four relevant evaluation scenarios arise. Cases \emph{(i)} and \emph{(ii)} represent situations where the \emph{Prediction Change} correctly ranks the local FA method, whereas \emph{(iii)} and \emph{(iv)} illustrate failure cases.} 
    \label{fig:dpc_cases}
\end{figure}

Consider the cases illustrated in Figure \ref{fig:dpc_cases}, where features are successively replaced with baseline values.
Without loss of generality, assume that increasing feature $x_i$ raises the model score $s_f^y(x)$.
Under this assumption, we obtain:
\begin{itemize}
    \item \emph{(i)} If the attribution predicts a \emph{decreasing model score} ($a_i<0$) and  $x_i$ \emph{decreases}, then $\mathrm{PC}^y_t(x)>0$. Hence, \textit{PC} correctly recognizes the FA method as erroneous. Case \emph{(ii)} is analogous for correctly identified features
    \item \emph{(iii)} If $a_i>0$ and $x_i$ \emph{decreases}, then $\mathrm{PC}^y_t(x)>0$, meaning that \textit{PC} incorrectly evaluates the FA as erroneous for this feature. Case \emph{(iv)} is the analogous incorrect recognition for the opposite case.
\end{itemize}
The \textit{PC} thus evaluates the FA method as correct only when the perturbation and attribution point in the same direction.
Because it relies solely on the change in model prediction, it cannot capture directional agreement.
This observation provides the intuition for our new metric: we extend \textit{PC} by incorporating information about the relative direction of the FA and the perturbation.

\subsection{Formal definition of DPC}
Mathematically, this directional information is fully represented by the signs of the perturbation and attribution.
If the FA and perturbation have different signs, the regular \textit{PC} must be inverted to yield a correct evaluation.
This can be achieved by multiplying the \textit{PC} with the signs of both the attribution and perturbation directions using the sign function $\sigma(\cdot)$.
The resulting \emph{Directed Prediction Change (DPC)} for perturbation step $t$ is defined as:

\begin{equation}
  \mathrm{DPC}^y_t(x) := \sigma\big(a_i(x)\big) \cdot \sigma(\pi_t(x)_i) \cdot \mathrm{PC}^y_t(x)  
\end{equation}

For perturbations affecting multiple features simultaneously (e.g., image data), we sum over all affected attributions and perturbations before computing their sign.

As in the case of \textit{PC}, faithful explanations cause larger prediction drops (lower \textit{DPC}) earlier or later depending on the evaluation order (MoRF or LeRF).
We again report the weighted \emph{Area Between Perturbation Curves} (ABPC) 
\cite{simicPerturbationEffectMetric2022} as the resulting metric. Thus higher values indicate better performance (\upar).

\section{Experiments and Results}
\label{sec:experiments}

\subsection{Setup}
\subsubsection{Datasets}
We employ two datasets from distinct modalities: a skin-lesion image dataset and a financial tabular dataset. 
All data samples are standardized, and both datasets are split into training, validation, and test sets using a $60{:}20{:}20$ ratio.

For the tabular data, we use the Home Equity Line of Credit (HELOC) dataset \cite{fico2018heloc}, which contains credit records labeled according to whether a loan was repaid or defaulted.
To avoid distortions in the explanations caused by imputation strategies, we exclude the three features with the highest proportion of missing values and remove all remaining data points with missing entries.
This yields $8\,290$ samples across $20$ features.
Models trained on this reduced dataset achieve performance comparable to those trained on the full version with imputation.

For the image data, we use the International Skin Imaging Collaboration (ISIC) challenge datasets covering the years 2016–2020 \cite{gutman2016isic, codella2017isic, codella2018isic, tschandl2019isic, rotemberg2020isic}.
The original task is multi-class classification of skin lesion types, which we simplify into a binary task distinguishing malignant from benign lesions.
Following \citet{cassidy2022analysis}, we merge the datasets across years, remove duplicates, and resize/crop all images to $224 \times 224$ pixels.
This results in $77\,227$ images with a strong class imbalance of approximately $80\%$ benign samples.

\subsubsection{Models}
We train two models per dataset: a simple linear model, offering higher interpretability, and a Deep Neural Network (DNN), representing a more complex nonlinear predictor.
For the tabular data, the DNN is a custom Multi-Layer Perceptron (MLP) with seven hidden layers. For image data we employ the \textit{InceptionV1} architecture \cite{szegedy2014going}.
All models use ReLU activations and are described in more detail in the Appendix.

\subsubsection{Explanation methods and metrics}
Following common practice, we use model logits as the score function $s_f^y$ for computing explanations and the predicted probabilities for evaluating explanation methods \cite{kokhlikyanCaptumUnifiedGeneric2020}.
This simplifies the FA task for the considered piecewise linear neural networks.

As discussed in Section \ref{sec:related}, all tested explanation methods expose a large number of hyperparameters, detailed in the Appendix.
To enable a comprehensive analysis of the evaluated metrics, we compare FA methods across a total of $4\,744$ hyperparameter configurations.
Unless otherwise noted, all experiments are performed on the validation splits to mimic a realistic hyperparameter-selection scenario for FA methods.

We employ the \textit{Infidelity} metric implementation from the \textit{Captum} framework \cite{kokhlikyanCaptumUnifiedGeneric2020}.
For local FA evaluation, we follow the approach by \citet{yeh2019infidelity}, where a subset of $k$ features is perturbed by adding Gaussian noise with $\sigma = 0.2$.

Guided Perturbation Experiments are implemented as successive replacement with the zero (mean) baseline.
Unless otherwise specified, we perform $20$ perturbation steps for Guided Perturbation Fidelity.
This ensures that on tabular data, each feature is perturbed separately (since HELOC contains 20 features), while on image data, a comparable number of model evaluations is performed.
We estimate \textit{Infidelity} by sampling $2^6 \cdot 20 = 1\,280$ perturbations on HELOC and $2^5 \cdot 20 = 640$ on ISIC.
The reduced sample count on ISIC is necessary due to the computational cost of \textit{Infidelity}. 

For all metrics, we employ a file-based cache for FA results, which substantially reduces computational demand and ensures that reported runtimes are independent of the explanation method used.
Nevertheless, due to computational constraints, we subsample the ISIC validation split to $3\,072$ samples (from $11\,487$) when evaluating \textit{Infidelity}.

\subsection{Results and Analysis}
\begin{figure}[t]
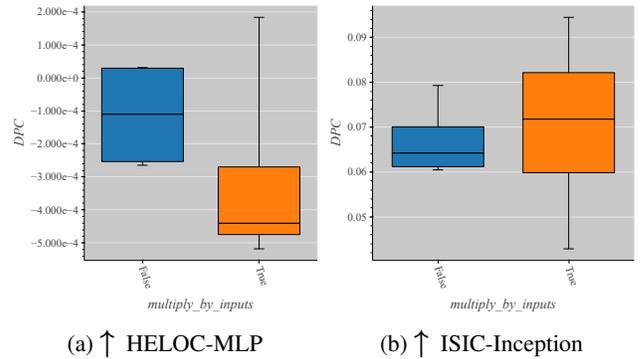

    \centering
    \begin{subfigure}{0.49\linewidth}
        \centering
        \includesvg[width=\linewidth]{images/HELOC-MLP_MBI-IG_DPC_0_clean.svg}
        \caption{\updir HELOC-MLP}
        \label{fig:mbi_dpc_nonlinear:heloc_ig}
    \end{subfigure}
    \begin{subfigure}{0.49\linewidth}
        \centering
        \includesvg[width=\linewidth]{images/ISIC-Inception_MBI-IG_DPC_0_clean.svg}
        \caption{\updir ISIC-Inception}
        \label{fig:mbi_dpc_nonlinear:isic_ig}
    \end{subfigure} 
    \caption{\textit{Directed Prediction Change} (\textit{DPC}) evaluation similar to Figure \ref{fig:mbi_pc}. \textit{DPC} improves the scoring of local FA methods, with a weaker effect on image data than on tabular data.}
    \label{fig:mbi_dpc_nonlinear}
\end{figure}

In general, evaluating the quality of FA metrics is challenging.
It is difficult to define a ground truth for the desired property and also difficult to measure it, even in controlled environments using synthetic datasets \cite{nauta2023anecdotal,dembinsky2025unifying}.
Furthermore, when using synthetic data, it is unclear to what extent the results transfer to real data \cite{dembinsky2025unifying}.

\subsubsection{\textit{DPC} prefers local FA methods} \label{sec:experiments:results:dpc_preference}

\begin{figure}[t]
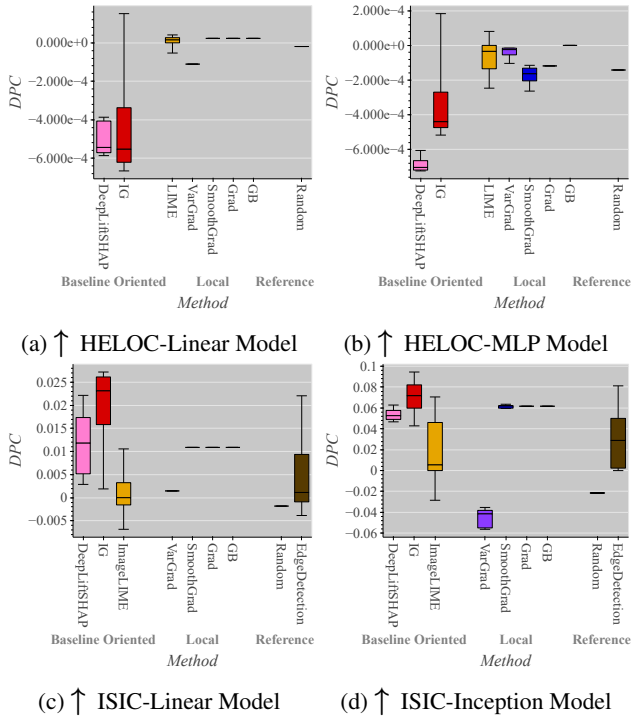

    \centering
    \begin{subfigure}{0.49\linewidth}
        \centering
        \includesvg[width=\linewidth]{images/HELOC-Linear_MBI=True_Method_DPC_0_clean.svg}
        \caption{\updir HELOC-Linear Model}
        \label{fig:method_mbi=true_dpc:heloc_linear}
    \end{subfigure}
    \begin{subfigure}{0.49\linewidth}
        \centering
        \includesvg[width=\linewidth]{images/HELOC-MLP_MBI=True_Method_DPC_0_clean.svg}
        \caption{\updir HELOC-MLP Model}
        \label{fig:method_mbi=true_dpc:heloc_mlp}
    \end{subfigure}\\
    \begin{subfigure}{0.49\linewidth}
        \centering
        \includesvg[width=\linewidth]{images/ISIC-Linear_MBI=True_Method_DPC_0_clean.svg}
        \caption{\updir ISIC-Linear Model}
        \label{fig:method_mbi=true_dpc:isic_linear}
    \end{subfigure}
    \begin{subfigure}{0.49\linewidth}
        \centering
        \includesvg[width=\linewidth]{images/ISIC-Inception_MBI=True_Method_DPC_0_clean.svg}
        \caption{\updir ISIC-Inception Model}
        \label{fig:method_mbi=true_dpc:isic_inception}
    \end{subfigure}\\
    \caption{Evaluation of FA methods according to the \textit{Directed Prediction Change} (\textit{DPC}). \textit{DPC} favors local FA methods compared to baseline-oriented methods.}
    \label{fig:method_mbi=true_dpc}
\end{figure}

First, we again resort to the experiments based on the approach by \citet{anconaGradientBasedAttributionMethods2019}, where baseline-oriented FA methods are converted into pseudo-local variants.
As shown in Figure \ref{fig:mbi_dpc_nonlinear}, for nonlinear models with \textit{Integrated Gradients}, \textit{DPC} consistently scores the pseudo-local variants higher (multiply\_by\_inputs $=False$).
Results for additional method and model combinations are provided in the Appendix.

On the ISIC dataset this effect is less pronounced.
Although \textit{DPC} increases the relative scores of pseudo-local FA methods compared to \textit{PC}, this increase can still leave relevant attributions ranked lower.
One possible reason is the limited number of perturbations for this modality, which introduces variance due to summation over many perturbed regions.
Another is the out-of-distribution issue described by \citet{hookerBenchmarkInterpretabilityMethods2019}, which may affect the perturbation experiment.
Hence, \textit{DPC} focuses more strongly on local FA than \textit{PC}, but both metrics may require additional work to mitigate out-of-distribution effects, which is out of scope for this paper.

We next inspect overall performance rankings obtained from \textit{DPC}.
Figure \ref{fig:method_mbi=true_dpc} compares the methods under consideration according to \textit{DPC}.
For the linear model, \textit{Gradient} is now evaluated substantially better than the other methods.
Especially on HELOC, \textit{DeepLiftSHAP} is often rated worse than a random attribution for the linear model.
This indicates that \textit{DPC} prefers more local methods than \textit{PC}. 

For the nonlinear models, the performances of individual FA methods overlap more strongly.
Across all dataset and model combinations, the random baseline is generally rated worse under \textit{DPC} than most FAs, which suggests that all analyzed methods encode some information about local model behavior.

\subsubsection{Comparing \textit{DPC} and local \textit{Infidelity}} \label{sec:experiments:results:dpc_vs_infidelity}

\begin{table}[t]
\centering
\begin{adjustbox}{width=0.5\linewidth}
 \begin{NiceTabular}{cc}
    	\toprule
    		\Block[c,c]{1-1}{Model} & \Block[c,c]{1-1}{Spearman Correlation}  \\\midrule %  & \Block[c,c]{1-1}{Pareto Size} & \Block[c,c]{1-1}{Total Size} 
    		\Block[c,l]{1-1}{HELOC-Linear} & \Block[c,r]{1-1}{$-0.71$}\\\cmidrule{2-2} %   & \Block[c,r]{1-1}{23} & \Block[c,r]{1-1}{44} 
    		\Block[c,l]{1-1}{HELOC-MLP} & \Block[c,r]{1-1}{$-0.42$}\\\midrule %    & \Block[c,r]{1-1}{17} & \Block[c,r]{1-1}{44}
    		\Block[c,l]{1-1}{ISIC-Linear} & \Block[c,r]{1-1}{$-0.57$} \\\cmidrule{2-2} %    & \Block[c,r]{1-1}{17} & \Block[c,r]{1-1}{44}
    		\Block[c,l]{1-1}{ISIC-Inception} & \Block[c,r]{1-1}{$-0.42$}\\ %  & \Block[c,r]{1-1}{6} & \Block[c,r]{1-1}{17} 
    	\bottomrule
    \end{NiceTabular}
   
\end{adjustbox}
    \caption{
        Spearman correlations between the overall validation scores of local \textit{Infidelity} and \textit{DPC} for all four models.
        All models exhibit significant anti-correlations.
    }
    \label{tab:corr_table_overall}
\end{table}

\begin{table}[t]
    \centering
     
    \begin{adjustbox}{width=\linewidth}
   
        \begin{NiceTabular}{cccc}
        	\toprule
        		\Block[c,c]{1-1}{Model} & \Block[c,c]{1-1}{Spearman Correlation} & \Block[c,c]{1-1}{Pareto Size} & \Block[c,c]{1-1}{Total Size} \\\midrule
        		 \Block[c,l]{1-1}{HELOC-Linear} & \Block[c,r]{1-1}{$-0.83$} & \Block[c,r]{1-1}{9} & \Block[c,r]{1-1}{729}\\\cmidrule{2-4}
        		\Block[c,l]{1-1}{HELOC-MLP} & \Block[c,r]{1-1}{$0.70$} & \Block[c,r]{1-1}{21} & \Block[c,r]{1-1}{729}\\\midrule
        		\Block[c,l]{1-1}{ISIC-Linear} & \Block[c,r]{1-1}{$-0.07$} & \Block[c,r]{1-1}{11} & \Block[c,r]{1-1}{1536}\\\cmidrule{2-4}
        		\Block[c,l]{1-1}{ISIC-Inception} & \Block[c,r]{1-1}{$-0.64$} & \Block[c,r]{1-1}{13} & \Block[c,r]{1-1}{1536}\\
        	\bottomrule
        \end{NiceTabular}
    \end{adjustbox}
   
    \caption{
        Spearman correlations between the overall validation scores of local \textit{Infidelity} and \textit{DPC} when evaluating only \textit{LIME}.
        All models show a small Pareto set relative to the total number of configurations, and, except for ISIC-Linear and HELOC-MLP, a strong anti-correlation as expected.
    } 
    \label{tab:corr_table_lime}
\end{table} 
We compare \textit{DPC} with local \textit{Infidelity}.
First, we analyze the correlations between the two metrics.
Since we test far more hyperparameter configurations for \textit{LIME} than for the other methods, we consider \textit{LIME} separately to avoid bias.
Table \ref{tab:corr_table_overall} shows that \textit{Infidelity} and \textit{DPC} exhibit significant anti-correlations according to the Spearman rank coefficient.
Because both metrics have opposing directions for optimal values, this suggests that they measure similar properties. 

A detailed examination of \textit{LIME} confirms this result.
We analyze the Pareto set \cite{ehrgottEfficiencyNondominance2005} for both metrics in Table \ref{tab:corr_table_lime}.
The Pareto set contains all configurations that are better in at least one metric and at least as good in the other than all other configurations.
The small Pareto sets relative to the total number of configurations indicate strong agreement between the metrics on optimal hyperparameters. 

\begin{figure}[t]
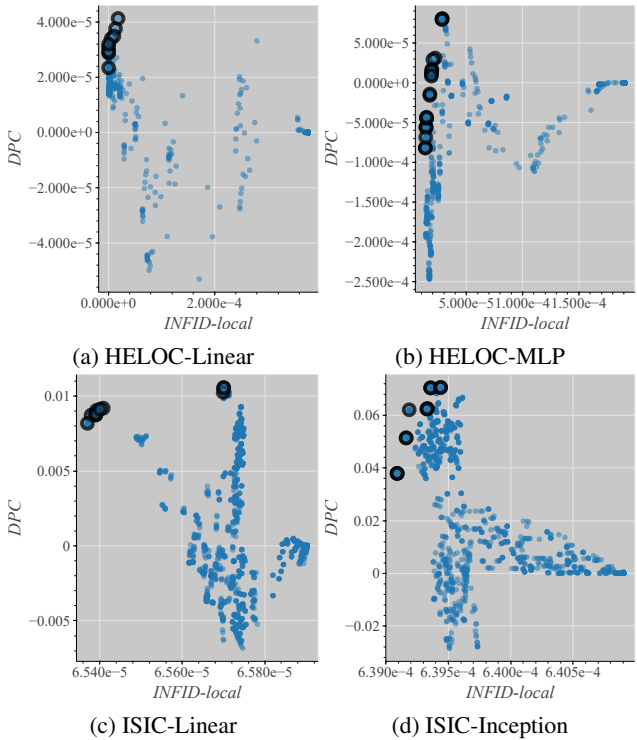

    \centering
    \begin{subfigure}{0.49\linewidth}
        \centering
        \includesvg[width=\linewidth]{images/faithfulness_scatter/LIME_HELOC-Linear_clean.svg}
        \caption{HELOC-Linear}
    \end{subfigure}
    \begin{subfigure}{0.49\linewidth}
        \centering
        \includesvg[width=\linewidth]{images/faithfulness_scatter/LIME_HELOC-MLP_clean.svg}
        \caption{HELOC-MLP}
    \end{subfigure}\\
    \begin{subfigure}{0.49\linewidth}
        \centering
        \includesvg[width=\linewidth]{images/faithfulness_scatter/ImageLIME_ISIC-Linear_clean.svg}
        \caption{ISIC-Linear}
    \end{subfigure}
    \begin{subfigure}{0.49\linewidth}
        \centering
        \includesvg[width=\linewidth]{images/faithfulness_scatter/ImageLIME_ISIC-Inception_clean.svg}
        \caption{ISIC-Inception}
    \end{subfigure}
    \caption{
        Scatter plots of local \textit{Infidelity} (x-axis) and \textit{DPC} (y-axis) across all tested  \textit{LIME} hyperparameters.
        Good configurations have low \textit{Infidelity} (\doar) and high \textit{DPC} (\upar).
        Pareto set elements are highlighted. 
    }
    \label{fig:lime_scatter}
\end{figure}

Correlations are generally stronger in absolute value for \textit{LIME} than for the model-wide aggregates excluding \textit{LIME}.
However, \textit{LIME} shows little correlation for the linear model on ISIC and a positive correlation for the MLP on HELOC. 

To investigate this more closely, we examine the scatter plots in Figure \ref{fig:lime_scatter}.
No simple monotonic relationships are evident, which explains why the Spearman coefficient can be misleading in this context and motivates a more nuanced analysis.

As a first observation, outlier hyperparameter configurations are identified for all four models.
These are characterized by being rated as the worst configurations according to \textit{Infidelity}, while \textit{DPC} assigns them scores near zero (indicating that even worse configurations exist under\textit{DPC}). 
Inspection of these cases reveals that the linear model trained within \textit{LIME} assigns near-zero importance to all features due to excessive $L_2$ regularization. 
In contrast to incorrect but seemingly plausible attributions, such degenerate explanations can be easily recognized by the users.
This work hence concludes that \textit{DPC} provides the more appropriate evaluation:
\textit{Infidelity} can rank erroneous explanations higher than configurations where the FA method clearly fails to produce a meaningful result, whereas \textit{DPC} pushes such non-informative configurations toward the bottom of the ranking. This avoids scenarios in which users might prefer incorrect yet plausible-looking explanations over implausible ones. 

For the ISIC-Linear model, we find additional outliers that receive high scores under \textit{Infidelity}.
These use the dataset mean as the pixel replacement value for \textit{LIME} perturbations.
Configurations ranked highly by \textit{DPC} (which receive medium scores under \textit{Infidelity}) instead use the mean of the explained image. 
Since the image mean is a more local perturbation than the dataset mean, we argue that these configurations better reflect local model behavior.
These outliers illustrate that \textit{Infidelity} can produce inaccurate evaluations relative to \textit{DPC}.

Overall, most \textit{Infidelity} scores lie in a very narrow range. 
In particular, for the HELOC-MLP model many configurations take nearly identical \textit{Infidelity} values.
Given the uncertainty observed in \textit{Infidelity} evaluations in Figure \ref{fig:intro_infidelity_confidence_and_runtime}, this lack of spread is problematic.

We suspect that this limited spread, together with the outliers, explains  the unexpected positive correlation for HELOC-MLP and the weak anti-correlation ISIC-Linear.
These effects support the effectiveness of \textit{DPC} relative  to \textit{Infidelity}, rather than indicating errors by our metric.

\subsubsection{Determinism and computational efficiency}
\label{sec:experiments:results:dpc_efficiency}
The computational efficiency of \textit{Infidelity} depends strongly on the number of samples. 
We therefore study the variance as a function of the sample count, based on the Monte Carlo approximation in equation \ref{eq:infid_main_equation}. 
Repeated evaluation with very large sample counts is infeasible when considering many FA methods and hyperparameters. 

We approximate the variance as a measure of the Monte Carlo uncertainty as follows.
First generate $N=640$ perturbations for ISIC are generated and the corresponding model scores are computed.
We then draw $64$ times $N_{\text{perturb}} \in \{20, 40, 80, 160, 320, 640\}$ perturbations from these $640$ and compute \textit{Infidelity} for each draw. 
The standard deviation across these evaluations estimates the uncertainty.
To summarize uncertainty at the method level, we average the aggregates (mean and standard deviation) across all hyperparameter settings of each method.
Note, that larger ratios $N_{\text{perturb}} / N$ induce greater overlap among sampled sets, which results in our procedure underestimating the true uncertainty.

Figure \ref{fig:intro_infidelity_confidence_and_runtime:uncertainty_image} shows the results.
Despite the aforementioned bias, the $\pm \sigma$ intervals show strong overlap with each other and their respective means.
It is therefore difficult to distinguish methods or to differentiate between hyperparameter configurations without using large numbers of model evaluations.

Our \textit{DPC} analysis uses $20$ perturbation steps\footnote{This corresponds to $40$ model evaluations because ABPC evaluates both LeRF and MoRF.}, which is significantly more efficient than \textit{Infidelity} where $640$ perturbations are required for a reliable evaluation.
We measure wall-clock runtime for both metrics when evaluating FAs of the Inception model on ISIC on a single A100-80GB GPU with the maximum feasible batch size.
The mean, median, and standard deviation across 20 runs are reported in Table \ref{fig:intro_infidelity_confidence_and_runtime:runtime_table}.
Even though our \textit{DPC} implementation does not employ the same batching optimizations as Captum's \textit{Infidelity}, we observe a mean speedup of $8.30$ and a median speedup of $9.91$.

Finally, \textit{PC} and \textit{DPC} can be computed from the same Guided Perturbation Experiment without additional model evaluations, since only the scoring differs.
By contrast, \textit{Infidelity} requires separate perturbation sets to evaluate both baseline-oriented and local behavior.
In scenarios where a holistic evaluation is desired, \textit{DPC} together with \textit{PC} yields an additional speedup factor of approximately $2$, for a total median speedup of approximately $20$ relative to \textit{Infidelity}.

\section{Conclusion}
\label{sec:discussion}
Motivated by the observation that existing metrics for evaluating local FA methods require many model evaluations due to Monte Carlo sampling, we developed an alternative metric that is both deterministic, and hence also trustworthy, and efficient.
To this end, we analyzed the Guided Perturbation Experiment and found that evaluation with \textit{Prediction Change} (\textit{PC}) is unsuitable for evaluating local FA methods.
Our main contribution is the \textit{Directed Prediction Change} (\textit{DPC}), which integrates the direction of the attribution and the applied perturbations into the evaluation within the Guided Perturbation Experiment. 

This modification enables effective evaluation of local FA methods and helps avoid error cases observed with the \textit{Infidelity} metric.
Furthermore, we achieve an almost tenfold median speedup over \textit{Infidelity}, because \textit{DPC} employs a deterministic procedure that does not require random sampling to obtain reliable results.
The efficient and trustworthy evaluation enabled by \textit{DPC} supports not only the final assessment of FA methods but also a holistic hyperparameter optimization, thereby contributing to the practical application of explainability methods in high-risk scenarios such as healthcare and finance.

\subsubsection{Limitations}
A limitation of our metric is reduced accuracy when measuring the performance of local FA methods on complex data types with many features, such as image data.
We hypothesize two causes.
First, summation during aggregation across multiple perturbation steps can combine conflicting perturbation directions.
Second, out-of-distribution effects may hamper the expressiveness for highly nonlinear models, even though our use of the weighted \textit{Area Between Perturbation Curves} (ABPC) already alleviates this issue. 

\subsubsection{Future Work}
To analyze and potentially mitigate or solve these issues, future work could study the effect of the number of perturbation steps on \textit{DPC} and explore adaptive choices for the perturbation step size.
It would further be valuable to examine local perturbations for \textit{DPC} instead of baseline replacement to counter out-of-distribution effects.
Our analysis therefore opens several promising directions in the rapidly evolving field of VXAI. 
Since these issues only slightly reduced the effectiveness of our metric, we conclude that \textit{DPC} is nonetheless a significant step toward an efficient and trustworthy evaluation of FA methods.

\section{Acknowledgements}
This research is funded by the German Federal Ministry for Digitalization and Government Modernization (BMDS) as part
of the project MISSION KI - Nationale Initiative für Künstliche Intelligenz und Datenökonomie with
the funding code 45KI22B021.

\bibliography{references}

\section{Appendix}
\subsection{Detailed Descriptions and Hyperparameters of the employed Explanation Algorithms}
\subsubsection{Vanilla Gradient}
The (Vanilla) Gradient method, introduced by Simonyan et al. \cite{simonyan2013deep}, explains a prediction by computing the sensitivity of the model output with respect to infinitesimal changes to the input features.
It is defined as:
\begin{equation}
    \mathbf{Grad}_f^y(x) = \nabla_x s_f^y(x).
\end{equation}
Features with larger gradients are assigned higher importance, as small changes in these features would cause larger changes in the prediction score.

\subsubsection{Guided Backpropagation}
The Guided Backpropagation (GB) method, introduced by Springenberg et al. \cite{springenberg2014striving}, extends the Vanilla Gradient method by modifying the backward pass through ReLU. 
Whereas the standard ReLU subgradient propagates gradients when the forward activation is positive, GB additionally suppresses negative upstream gradients. 
Formally, let $g(x)$ denote the pre-activation input to a ReLU, and let $R\;=\; \frac{\partial s_f^y(x)}{\partial \,\mathrm{ReLU}\big(g(x)\big)}$ be the upstream gradient at that node.
GB replaces the standard ReLU derivative with
\begin{equation}
    \frac{\partial s_f^y(x)}{\partial g(x)} \;:=\; R  \cdot \mathds{1}[g(x)>0] \cdot\mathds{1}[R>0],
\end{equation}
i.e., the gradient is zeroed whenever the forward activation is non-positive or the incoming gradient is non-positive.
Originally proposed as a visualization heuristic \cite{springenberg2014striving}, GB has been criticized for producing edge-detector-like outputs and for limited class sensitivity \cite{nie2018theoretical}.

\subsubsection{Integrated Gradient}
Sundararajan et al. propose the Integrated Gradients (IG) method as an attribution method that satisfies a set of desirable theoretical properties \cite{sundararajan2017axiomatic}.
Rather than computing a single gradient, it computes attributions by integrating gradients along a path from a given baseline $x_0$ to the input:
\begin{equation}
    \mathbf{IG}_{f, x_0}^y(x) = (x-x_0) \int_0^1 \nabla_x s_f^y\big(x_0 + \alpha(x-x_0)\big)\mathrm{d}\alpha .
\end{equation}
Sundararajan et al. show that this method distributes the difference in model score between the baseline and the input among the considered features, thereby providing a baseline-oriented FA method.

\subsubsection{SmoothGrad \& VarGrad}
A strategy to reduce artifacts in noisy FAs is presented by Smilkov et al. with SmoothGrad (SG) \cite{smilkov2017smoothgrad}. 
They extend (in principle) any FA method by averaging the attributions over multiple small perturbations of the input.
For simplicity, we implement their method by wrapping the Gradient method and employ Gaussian noise for generating perturbations, as in Smilkov et al., resulting in the explanation method:
\begin{equation}
    \mathbf{SG}_f^y(x) = \ex_{X\sim\mathcal{N}(x,\sigma I)}\!\big[ \nabla_X s_f^y(X) \big].
\end{equation}

A variance-based analog to SmoothGrad is presented by Adebayo et al. \cite{adebayo2018local} in the form of VarGrad:
\begin{equation}
    \mathbf{VG}_f^y(x) = \var_{X\sim\mathcal{N}(x,\sigma I)}\!\big[ \nabla_X s_f^y(X) \big].
\end{equation}

While SmoothGrad reduces noise through averaging, VarGrad highlights high variance across perturbations, thereby estimating higher-order derivatives of the model score for the considered class \cite{seoNoiseaddingMethodsSaliency2018}.

\subsubsection{LIME}
The Local Interpretable Model-Agnostic Explanations (LIME) method was introduced by Ribeiro et al. \cite{ribeiro2016why} and approximates local model behavior without relying on the model's gradient.
It extracts attributions from an inherently interpretable model $g$ with complexity $\Omega(g)$, which is trained through loss $\mathcal{L}$ to mimic the predictive behavior of $f$ in a neighborhood around the input $x$.:
\begin{equation}
    \mathbf{LIME}_f^y(x) = \arg \min_{g\in G} \;\mathcal{L}^y(f,g,x) + \Omega(g).
\end{equation}

Similar to Ribeiro et al., we train a linear regression model as the interpretable model $g$ and use its weights as attribution scores. 

While this approach does not require the model to be differentiable, unlike the previously discussed methods, it introduces several parameters.
First, the perturbation scheme and the number of perturbed samples must be defined. 
Second, a complexity measure has to be chosen.
Finally, perturbations that are too far from the data point may no longer reflect local model behavior, which is why Ribeiro et al. propose a weighting function for distant samples \cite{ribeiro2016why}.
We account for all these parameters in our experiments.

\subsubsection{SHAP}
The SHapley Additive exPlanations (SHAP) method, introduced by Lundberg and Lee \cite{lundberg2017shap}, calculates attributions w.r.t. the average model-prediction baseline by leveraging cooperative game theory.
Each feature is interpreted as a player in a cooperative game, and the payout (i.e., the difference in model prediction compared to the average model prediction) is distributed fairly among players.
This is achieved using Shapley Values \cite{shapley1953value}; for details on how SHAP computes the average marginal contribution of a given feature across all feature subsets, we refer to Lundberg and Lee  \cite{lundberg2017shap}.

Although SHAP satisfies desirable axiomatic properties \cite{lundberg2017shap, sundararajanManyShapleyValues2020}, in practice an exact computation is intractable for more than a handful of features.
In this work we therefore use DeepLiftSHAP \cite{lundberg2017shap}, a model-specific implementation that combines the DeepLIFT modified backpropagation procedure \cite{shrikumar2017learning} with Shapley-value-based weighting to approximate SHAP values efficiently for deep neural networks.

\subsubsection{Model-agnostic baselines}
\paragraph{Random attributions}
Random explanations act as data-independent references that any meaningful explanation algorithm should outperform. 
We implement two variants: 
(1) a constant random attribution \textbf{RndC}, where a random vector is sampled once and used for all inputs; 
(2) a non-constant variant \textbf{RndNC}, where a new random attribution is sampled for each input: 
\begin{align}
    \mathbf{RndC}_f^y(x) = r,\quad r \sim \mathcal{N}(0,I), \\
    \mathbf{RndNC}_f^y(x) \sim \mathcal{N}(0,I).
\end{align}

\paragraph{Edge detection}
Since some FA methods behave similarly to edge detectors on image data \cite{adebayo2018sanity}, we also include edge detection as a data-dependent na\"ive explanation. 
Specifically, we use two established computer vision techniques: (1) the Sobel gradient filter \cite{sobel2014isotropic} and (2) the Canny edge detection algorithm \cite{canny1986computational}. 
To reduce sparsity and increase object coverage, we apply a Gaussian filter as a post-processing step and vary the filter width as a hyperparameter.

\subsubsection{Hyperparameters}
We explore a large variety of hyperparameters across all FA methods. This includes, on the one hand, different baseline choices for Integrated Gradients and artificial biasing toward either class for the reference distribution of DeepLiftSHAP, and, on the other hand, the large variety of hyperparameters used by LIME as well as different choices for the noise used by SmoothGrad and VarGrad.

For all methods, except for the individual LIME approaches and EdgeDetection, two models on two datasets are considered (for the individual LIME approaches and EdgeDetection, two models on one dataset each are considered). This amounts to an analysis of a total of $4\,744$ setups on the corresponding validation splits, presented in Table \ref{table:tested_feature_attribution_params}.

\begin{table*}
	\centering
	
	\begin{adjustbox}{width=0.95\textwidth}
		\begin{NiceTabular}{m{0.15\textwidth}m{0.15\textwidth}p{0.8\textwidth}m{0.1\textwidth}}%
			\toprule
			\Block[c,c]{1-1}{Method Type} & \Block[c,c]{1-1}{Method} & \Block[c,c]{1-1}{Tested Hyperparameters} & \Block[c,c]{1-1}{\# Configurations} \\
			\midrule
			\Block[c,c]{3-1}{Propagation Based} & \Block[c,c]{1-1}{Gradient} & \Block[c,l]{1-1}{$-$} & \Block[c,c]{1-1}{$1$} \\ \cmidrule{2-4}
			& \Block[c,c]{1-1}{GB} &  \Block[c,l]{1-1}{$-$} & \Block[c,c]{1-1}{$1$} \\ \cmidrule{2-4}
			& \Block[c,c]{1-1}{IG} & \Block[c,l]{1-1}{min, mean (0), median, and max baselines\\ Input multiplication (True or False)\\ 64 samples along a straight path} & \Block[c,c]{1-1}{$8$}\\ \midrule
			\Block[c,c]{3-1}{Perturbation Based} & \Block[c,c]{1-1}{(Tabular) LIME} & \Block[c,l]{1-1}{$\alpha \in \{0.000055, 0.0001, 0.00055, 0.001, 0.0055, 0.01, 0.055, 0.1, 0.55\}$\\$\sigma_k \in \{0.25, 0.375, 0.5, 0.625, 0.75, 0.875, 1.0, 1.125, 1.25\}$\\$\sigma_s \in \{0.1, 0.5, 1\}$\\ $n_{\text{samples}} \in \{64, 256, 1024\}$} & \Block[c,c]{1-1}{$729$} \\ \cmidrule{2-4} 
			& \Block[c,c]{1-1}{(Image) LIME} & \Block[c,l]{1-1}{$\alpha \in \{0.00055, 0.001, 0.0055, 0.01, 0.055, 0.1, 0.55, 1\}$\\ $\sigma_k \in \{0.125, 0.25, 0.375, 0.5\}$\\ $n_{\text{samples}} = 1024$\\ Replacement value: Segment-Mean, Image-Mean, Dataset-Mean\\ Segmentation algorithm: Quickshift or SLIC\\Seg. preprocess $\sigma \in \{0, 4\}$ \\ Seg. Quickshift: max\_dist$\in \{5, 7.5, 10, 200\}$ \\ Seg. SLIC: $n_{\text{segments}} \in \{48, 64, 96, 128\}$}& \Block[c,c]{1-1}{$1536$} \\ \cmidrule{2-4} 
			& \Block[c,c]{1-1}{DeepLiftSHAP} &  \Block[c,l]{1-1}{Stratified baseline distribution with expected label $y \in \{0.0, 0.1, 0.2, 0.3, 0.4, 0.5, 0.6, 0.7, 0.8, 0.9, 1.0\}$ or Random\\Input multiplication (True or False)\\$n_{\text{samples}} = 1024$ for tabular data\\$n_{\text{samples}} = 128$ for image data} & \Block[c,c]{1-1}{$24$}  \\ \midrule
			\Block[c,c]{2-1}{Wrapping Approaches} & \Block[c,c]{1-1}{SmoothGrad} & \Block[c,l]{1-1}{$\sigma \in \{0.01, 0.1, 0.25, 0.5, 1\}$} & \Block[c,c]{1-1}{$5$} \\ \cmidrule{2-4} 
			& \Block[c,c]{1-1}{VarGrad} & \Block[c,l]{1-1}{$\sigma \in \{0.01, 0.1, 0.25, 0.5, 1\}$} & \Block[c,c]{1-1}{$5$} \\ 
			\midrule
			\Block[c,c]{2-1}{(Model-Agnostic) Baselines}& \Block[c,c]{1-1}{Random Attribution} & \Block[c,l]{1-1}{Use a constant value (True or False)} & \Block[c,c]{1-1}{$2$} \\ \cmidrule{2-4}
			& \Block[c,c]{1-1}{Edge Detection} & \Block[c,l]{1-1}{Postprocess $\sigma_{\text{post}} \in \{0.0, 2.0, 4.0, 8.0\}$\\ Sobel or Canny edge-detection algorithm\\ Canny smoothing $\sigma_{\text{smooth}} \in \{1.0, 2.0, 4.0\}$} & \Block[c,c]{1-1}{$16$} \\ 
			\bottomrule
		\end{NiceTabular}
	\end{adjustbox}
	\caption[Tested hyperparameter configurations for all FA methods]{List of all tested hyperparameter configurations for all FA methods explored in this work.}
	\label{table:tested_feature_attribution_params}
\end{table*}

\subsection{Proof of Theorem 1}
Let $\mathcal{A}$ be a FA method, $x$ be the considered data point, $x' \in \Rd$ be a baseline, and $s_f^y$ be a scoring function for a model $f$ and a class $y$. Further, let $\mathcal{A}$ fulfill Sensitivity-$N$ with the baseline $x'$ for all $x'' \in \R^d$ on all perturbation paths between $x$ and $x'$ (inclusive) for all $1 \leq N \leq d$. 

Let $x'$ denote the baseline used by the Guided Perturbation Experiment, $X:=\{1, \dots, d\}$, and 
\begin{equation}
    \tau \in X^d: \forall i, j \in X: i \neq j \Leftrightarrow \tau_i \neq \tau_j
\end{equation}

We define the perturbation function $\pi_t(x) = \pi_{t-1}(x_{x \setminus \{\tau_t\}};x')$, where $x_{\{\tau_t\}};x'$ denotes the replacement of the feature $\tau_t$ with the corresponding feature of $x'$. We observe that $\pi$ encompasses all valid perturbation functions used by the Guided Perturbation Experiment.

Next, we investigate an arbitrary set $S \subseteq X$,$|S| \leq d$. By the definition of Sensitivity-$N$ and by using a telescoping sum, we obtain:
\begin{align} \label{eq:sensitivity_N_is_pc}
    \sum_{i \in S} \mathcal{A}_{f}^y(x)_i =& s_f(x) - s_f(x_{X \setminus S};x') \\
    =& s_f(x) - s_f(\pi_{|S|}(x))  \\
     =&  \sum_{i=1}^{|S|} s_f^{y}\big(\pi_{i-1}(x)\big) - s_f^{y}\big(\pi_i(x)\big)\\ %s_f(x) - s_f(\pi_{1}(x)) + s_f(\pi_{1}(x))  - \dots \\& - s_f(\pi_{N - 1}(x)) + s_f(\pi_{N - 1}(x)) - s_f(\pi_{N}(x))   \notag\\
    =& \sum_{i=1}^{|S|} - \mathrm{PC}^y_i(x)
\end{align}

We utilize this result to show the theorem by induction over the performed perturbation steps. 

Since we can choose $\tau_1$ freely, we obtain for the first perturbation step:
\begin{equation} 
    \mathcal{A}_{f}^y(x)_i = - \mathrm{PC}^y_i(x) \forall i \in \{1, \dots, d\}
\end{equation}
Hence, choosing the feature to perturb based on the largest attribution value will choose the feature causing the largest drop in prediction.

Now for any $\tau_i$, $i > 1$ we first observe that $\pi_{i-1}(x)$ is again a data point for which our previous analysis applies, and thus the attribution value of that data point is indicative of the feature causing the largest drop in prediction.
Since, by assumption, the order of features by the FA method stays consistent along the perturbation path, we have that for all steps up to $i$ the Prediction Change is maximal. 

\qed

\subsection{Models}
For the HELOC dataset, the nonlinear model is a custom fully-connected neural network with sufficient capacity to achieve $99.9\%$ training accuracy \footnote{
    The network has hidden dimensions $[32, 128, 256, 128, 256, 128, 32]$ with batch-normalization \cite{ioffe2015batch} and dropout \cite{srivastava2014dropout}.
}.
For the ISIC dataset, we employ the \textit{InceptionV1} architecture \cite{szegedy2014going} as a representative deep model.
Both networks use ReLU activations.

To improve model performance, we apply suitable data augmentation strategies.
On the HELOC dataset, we add Gaussian noise to the features.
On the ISIC dataset, we follow the augmentation scheme of the winning submission to the ISIC 2020 Challenge  \cite{ha2020identifying}.

\begin{table}
	\centering
		\begin{NiceTabular}{ccc}
			\toprule
			\Block[c,c]{1-1}{Model} & \Block[c,c]{1-1}{AUROC} & \Block[c,c]{1-1}{Accuracy} \\\midrule
			\Block[c,l]{1-1}{ISIC-Inception} & \Block[c,c]{1-1}{$92.91\%$} & \Block[c,c]{1-1}{$88.34\%$} \\\cmidrule{2-3}
			\Block[c,l]{1-1}{ISIC-Linear} &  \Block[c,c]{1-1}{$82.99\%$} & \Block[c,c]{1-1}{$69.29\%$} \\\midrule
			\Block[c,l]{1-1}{HELOC-MLP} &  \Block[c,c]{1-1}{$79.72\%$} & \Block[c,c]{1-1}{$73.01\%$} \\\cmidrule{2-3}
			\Block[c,l]{1-1}{HELOC-Linear}  & \Block[c,c]{1-1}{$79.46\%$} & \Block[c,c]{1-1}{$73.43\%$} \\
			\bottomrule
		\end{NiceTabular}
        \vspace{0.25cm}
	\caption[Performance of the models trained on both datasets]{Performance of all trained models.}
	\label{table:model_results}
\end{table}

All models are optimized on the training split of their respective datasets using a logistic regression loss and the AdamW optimizer with \textit{amsgrad} \cite{kingmaAdamMethodStochastic2017,reddiConvergenceAdam2019, loshchilovDecoupledWeightDecay2019}.
We performed extensive hyperparameter tuning, ensuring that all models achieved competitive performance on their respective test splits, as shown in Table \ref{table:model_results}. 

\subsection{Additional input-multiplication experiments}
\begin{figure}
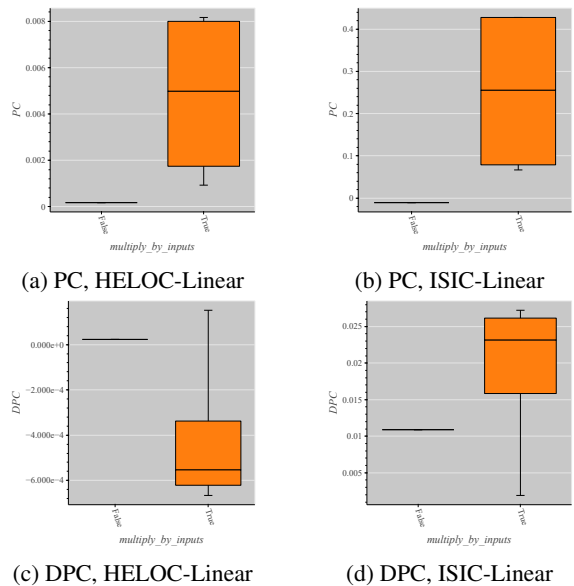

    \centering
    
    \begin{subfigure}{0.49\linewidth}
        \centering
        \includesvg[width=0.8\linewidth]{images/HELOC-Linear_MBI-IG_PC_0_clean.svg}
        \caption{PC, HELOC-Linear}
    \end{subfigure}
    \begin{subfigure}{0.49\linewidth}
        \centering
        \includesvg[width=0.8\linewidth]{images/ISIC-Linear_MBI-IG_PC_0_clean.svg}
        \caption{PC, ISIC-Linear}
    \end{subfigure}\\
    \begin{subfigure}{0.49\linewidth}
        \centering
        \includesvg[width=0.8\linewidth]{images/HELOC-Linear_MBI-IG_DPC_0_clean.svg}
        \caption{DPC, HELOC-Linear}
        \label{fig:mbi_dpc_linear:heloc_ig}
    \end{subfigure}
    \begin{subfigure}{0.49\linewidth}
        \centering
        \includesvg[width=0.8\linewidth]{images/ISIC-Linear_MBI-IG_DPC_0_clean.svg}
        \caption{DPC, ISIC-Linear}
        \label{fig:mbi_dpc_linear:isic_ig}
    \end{subfigure} 
    \caption{Directed Prediction Change (DPC) and Prediction Change (PC) evaluation as in Figure \ref{fig:mbi_pc} for Integrated Gradients on linear models.}
    \label{fig:mbi_ig_linear}
\end{figure}

We provide extended results when converting baseline-oriented FA methods into pseudo-local approaches as described in Section \ref{sec:new_metric:pc_analysis}. Figure \ref{fig:mbi_ig_linear} shows the results of Integrated Gradients for both metrics and the employed linear models. Similarly, Figure \ref{fig:mbi_deepliftshap_linear} and Figure \ref{fig:mbi_deepliftshap_nonlinear} show the results when investigating \textit{DeepLiftSHAP}. We observe in these extended evaluations the same patterns described in the main paper:
\textit{PC} prefers baseline-oriented approaches, whereas \textit{DPC} more strongly prefers local FA methods. We again find that the effect of our modification is weaker on image data but still visible.

\begin{figure}
    \centering
    \begin{subfigure}{0.49\linewidth}
        \centering
        \includesvg[width=\linewidth]{images/HELOC-Linear_MBI-DeepSHAP_PC_0_clean.svg}
        \caption{PC, HELOC-Linear}
    \end{subfigure}
    \begin{subfigure}{0.49\linewidth}
        \centering
        \includesvg[width=\linewidth]{images/ISIC-Linear_MBI-DeepSHAP_PC_0_clean.svg}
        \caption{PC, ISIC-Linear}
    \end{subfigure}\\
    \begin{subfigure}{0.49\linewidth}
        \centering
        \includesvg[width=\linewidth]{images/HELOC-Linear_MBI-DeepSHAP_DPC_0_clean.svg}
        \caption{DPC, HELOC-Linear}
        \label{fig:mbi_dpc_linear:heloc_deepliftshap}
    \end{subfigure}
    \begin{subfigure}{0.49\linewidth}
        \centering
        \includesvg[width=\linewidth]{images/ISIC-Linear_MBI-DeepSHAP_DPC_0_clean.svg}
        \caption{DPC, ISIC-Linear}
        \label{fig:mbi_dpc_linear:isic_deepliftshap}
    \end{subfigure}
     
    \caption{Directed Prediction Change (DPC) and Prediction Change (PC) evaluation as in Figure \ref{fig:mbi_pc} for DeepLiftSHAP on linear models.}
    \label{fig:mbi_deepliftshap_linear}
\end{figure}
\begin{figure}
    \centering
    \begin{subfigure}{0.49\linewidth}
        \centering
        \includesvg[width=\linewidth]{images/HELOC-MLP_MBI-DeepSHAP_PC_0_clean.svg}
        \caption{PC, HELOC-MLP}
    \end{subfigure}
    \begin{subfigure}{0.49\linewidth}
        \centering
        \includesvg[width=\linewidth]{images/ISIC-Inception_MBI-DeepSHAP_PC_0_clean.svg}
        \caption{PC, ISIC-Inception}
    \end{subfigure}\\
    \begin{subfigure}{0.49\linewidth}
        \centering
        \includesvg[width=\linewidth]{images/HELOC-MLP_MBI-DeepSHAP_DPC_0_clean.svg}
        \caption{DPC, HELOC-MLP}
        \label{fig:mbi_dpc_nonlinear:heloc_deepliftshap}
    \end{subfigure}
    \begin{subfigure}{0.49\linewidth}
        \centering
        \includesvg[width=\linewidth]{images/ISIC-Inception_MBI-DeepSHAP_DPC_0_clean.svg}
        \caption{DPC, ISIC-Inception}
        \label{fig:mbi_dpc_nonlinear:isic_deepliftshap}
    \end{subfigure}
     
    \caption{Directed Prediction Change (DPC) and Prediction Change (PC) evaluation as in Figure \ref{fig:mbi_pc} for DeepLiftSHAP on nonlinear models.}
    \label{fig:mbi_deepliftshap_nonlinear}
\end{figure}
\end{document}